\documentclass[letterpaper, 10 pt, conference]{ieeeconf}  

\IEEEoverridecommandlockouts                              

\usepackage{graphicx} 
\usepackage{multirow}
\usepackage{array}
\usepackage{amsmath}
 \usepackage{booktabs}
 \usepackage{xcolor}

\usepackage{hyperref}
\hypersetup{
    colorlinks=true,
    linkcolor=black,
    filecolor=magenta,      
    urlcolor=blue,
}

\title{\LARGE \bf
\textit{FLYNN}: Robust Neural Network for Robot Navigation using Fly Brain Topology
}

\author{Benquan Wang$^{1}$ and Jingdao Chen$^{2}$
\thanks{$^{1}$Benquan Wang is with the Department of Computer Science Engineering, Mississippi State University, Mississippi State, MS 39762, USA
        {\tt\small bw1918@msstate.edu}}%
\thanks{$^{2}$Jingdao Chen is with the Department of Computer Science Engineering, Mississippi State University, Mississippi State, MS 39762, USA
        {\tt\small chenjingdao@cse.msstate.edu}}%
\thanks{$^{*}$Source code is available at \url{github.com/ben-gitdev/fly-gym}}%
}

\begin{document}

\maketitle
\thispagestyle{empty}
\pagestyle{empty}

\begin{abstract}

While deep learning models achieve state-of-the-art performance in complex tasks, they remain brittle when faced with new environments or sensory deprivation. In contrast, biological systems exhibit remarkable tolerance to these challenges. We address this vulnerability by developing a recurrent neural network (RNN) whose architecture is directly derived from the synaptic-resolution brain connectome of the fruit fly \textit{Drosophila melanogaster}. We demonstrate the feasibility of training the fly connectome neural network (FLYNN) to perform vision-based navigation in MuJoCo, achieving performance comparable to modern hand-crafted networks of similar parameter counts. Crucially, FLYNN exhibits superior resistance to out-of-distribution (OOD) data and tolerance to sensory loss without further training. It remained functional even under total vision loss while hand-crafted networks largely failed, even when specifically trained with camera dropout. Principal Component Analysis (PCA) of the internal state of FLYNN suggests that it exhibits a particularly high degree of representational modularity, which might be related to its robustness. Our work provides a new direction for designing resilient artificial agents following the topology of biological brains. 

\end{abstract}

\section{INTRODUCTION}

Artificial Neural Networks (ANNs) originated from biologically inspired artificial neurons \cite{perceptron1958}, and the biological brain remains a foundational source of inspiration even as modern architectures are primarily hand-crafted. The Convolutional Neural Network (CNN) \cite{CNN_AlexNet} is a prominent example, followed by Recurrent Neural Networks (RNNs) \cite{bptt}, Long Short-Term Memory (LSTM) units \cite{lstm}, and Transformers \cite{attention}. State-of-the-art ANNs now frequently outperform humans on a wide range of complex tasks.

Navigation is one of the areas in which robustness is critical. While current state-of-the-art ANN models perform well under ideal conditions, they are known to be vulnerable to  out-of-distribution (OOD) data \cite{shoeb2025out} and sensor information degradation or deprivation. In autonomous driving, for example, a never-before-seen situation or the loss of a camera stream typically triggers an emergency handover, as the models rely on continuous integration of all inputs \cite{ceccarelli2022rgb}. 

In contrast, biological organisms are extraordinarily robust to these failure modes. For example, an animal that loses an eye typically adapts quickly and continues to navigate. Such resilience is likely rooted in the distinctive architecture of the biological brain. In \textit{Drosophila}, for instance, the visual system is composed of distinct, highly organized neuropils -- including the lamina, medulla, lobula plate, and lobula -- specialized for calculating optical flow from visual input \cite{drosophila_visual_system,how_flys_see_motion}. Furthermore, the central complex of the \textit{Drosophila} brain contains a population of ring attractor neurons that serve as a heading direction calculator for navigation \cite{ring_neuron_2020,ring_neuron_2022}. These well-organized structures serve as a structural prior that enables the organism to maintain stable internal representations of space even in unknown environments or under sensory deprivation.

Extensive studies have focused on improving OOD generalization of ANN models for robust navigation. Efforts include designing specialized training strategies like domain randomization \cite{domain_rand_zhang2024}, and developing new network architectures like liquid neural networks \cite{liquidnet_suresh2025liquid,liqid_net}. Large language models have also been explored for robust navigation tasks \cite{llm_nav_aasi2025generating,llm_nav_qiao2025open}.
Many studies have also focused on enhancing model tolerance to image degradation, occlusion, and total sensor loss \cite{input_corruption_miscov,input_occlusion,input_sensor_failure}. However, maintaining navigational robustness in new environments or during camera failure remains a significant challenge. The extraordinary resilience exhibited by biological organisms, combined with the recent availability of the complete \textit{Drosophila} brain connectome \cite{fly_connectome_1,fly_connectome_2} -- the first connectome of an animal with a complex vision system -- presents a unique opportunity to develop a biologically faithful architecture specifically designed to address these challenges. To our knowledge, this is the first study to exploit the \textit{Drosophila} connectome in the context of robot navigation tasks. Most existing works have focused on using biologically realistic simulations to investigate and understand the mechanisms of different systems of the brain, for example, the visual system \cite{2024Lappalainen} and the taste system \cite{2024Shiu,drosophila_taste_sys_sim}.

In this work, we translated the synaptic-resolution \textit{Drosophila} connectome into an RNN. By constraining the network's connectivity to this biological blueprint, we inherited the functional redundancy and modularity characteristic of biological systems. We evaluated the fly connectome neural network (FLYNN) in a simulated MuJoCo environment \cite{2012MuJoCo}, compared it against standard networks of similar parameter sizes such as EfficientNet-B0 \cite{efficientnet}, MobileNet-large-v3 \cite{mobilenet}, and small-world RNN \cite{small_world}, and investigated the sources of robustness in biological neural architectures.

Our results demonstrate three key findings:
\begin{enumerate}
    \item Feasibility: An RNN architecture constrained strictly by biological connectivity can be trained to perform complex tasks, such as vision-based multi-sensory navigation.
    \item Competitive Performance: The FLYNN achieved performance levels comparable to modern hand-crafted models of similar parameter counts.
    \item Graceful Degradation: Most notably, and as observed in biological organisms, the FLYNN exhibits unique robustness to OOD data and sensor ablation even though it was never trained for this.
\end{enumerate}

\section{Methods}

\begin{figure}
    \centering
    \includegraphics[width=0.85\linewidth]{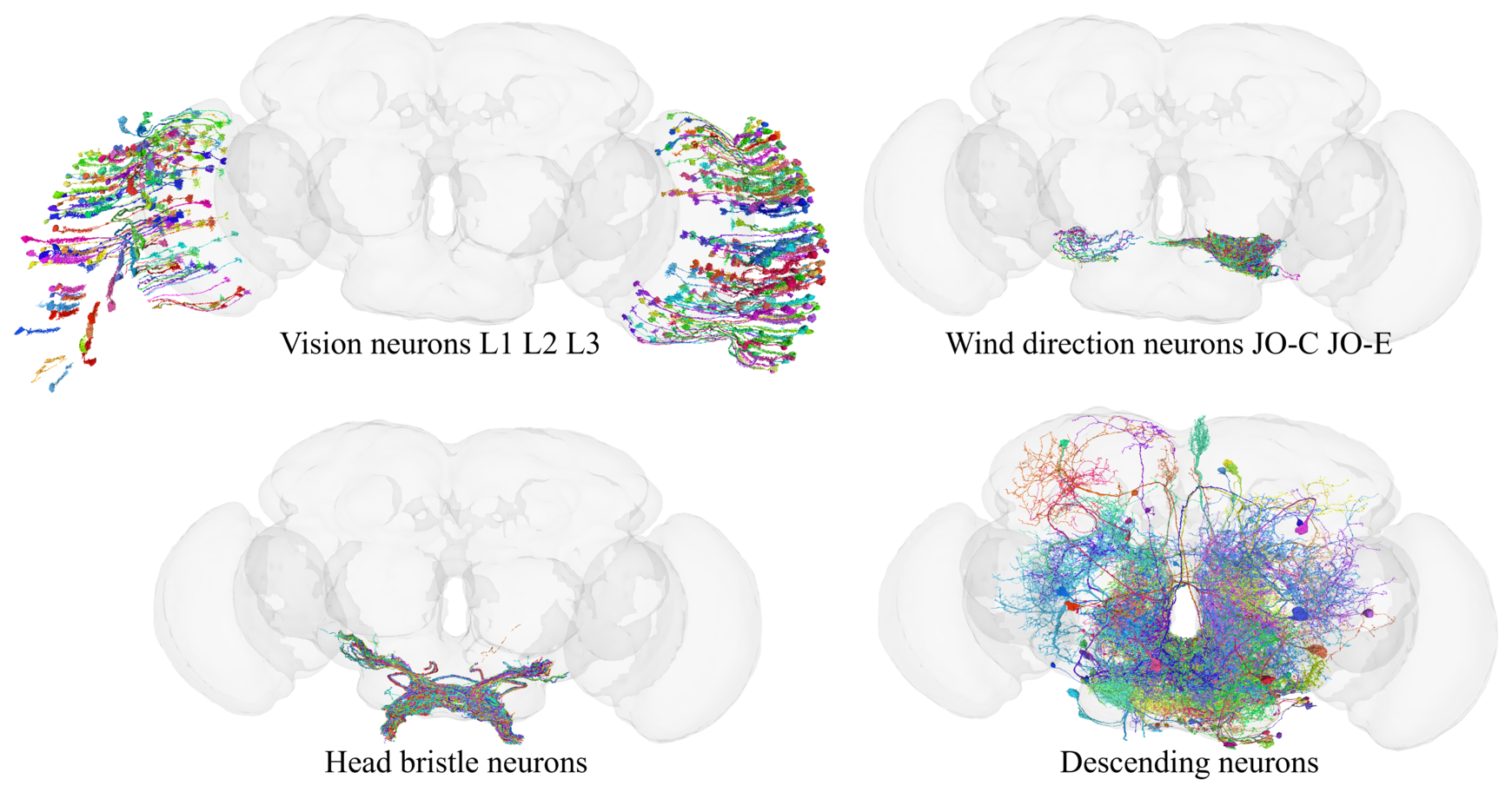}
    \caption{Subsets of sensory neurons and descending neurons used in this work in the FAFB v783 connectome dataset. Credit: FlyWire.ai.}
    \label{fig:neurons}
\end{figure}

\begin{figure}
    \centering
    \includegraphics[width=0.48\textwidth]{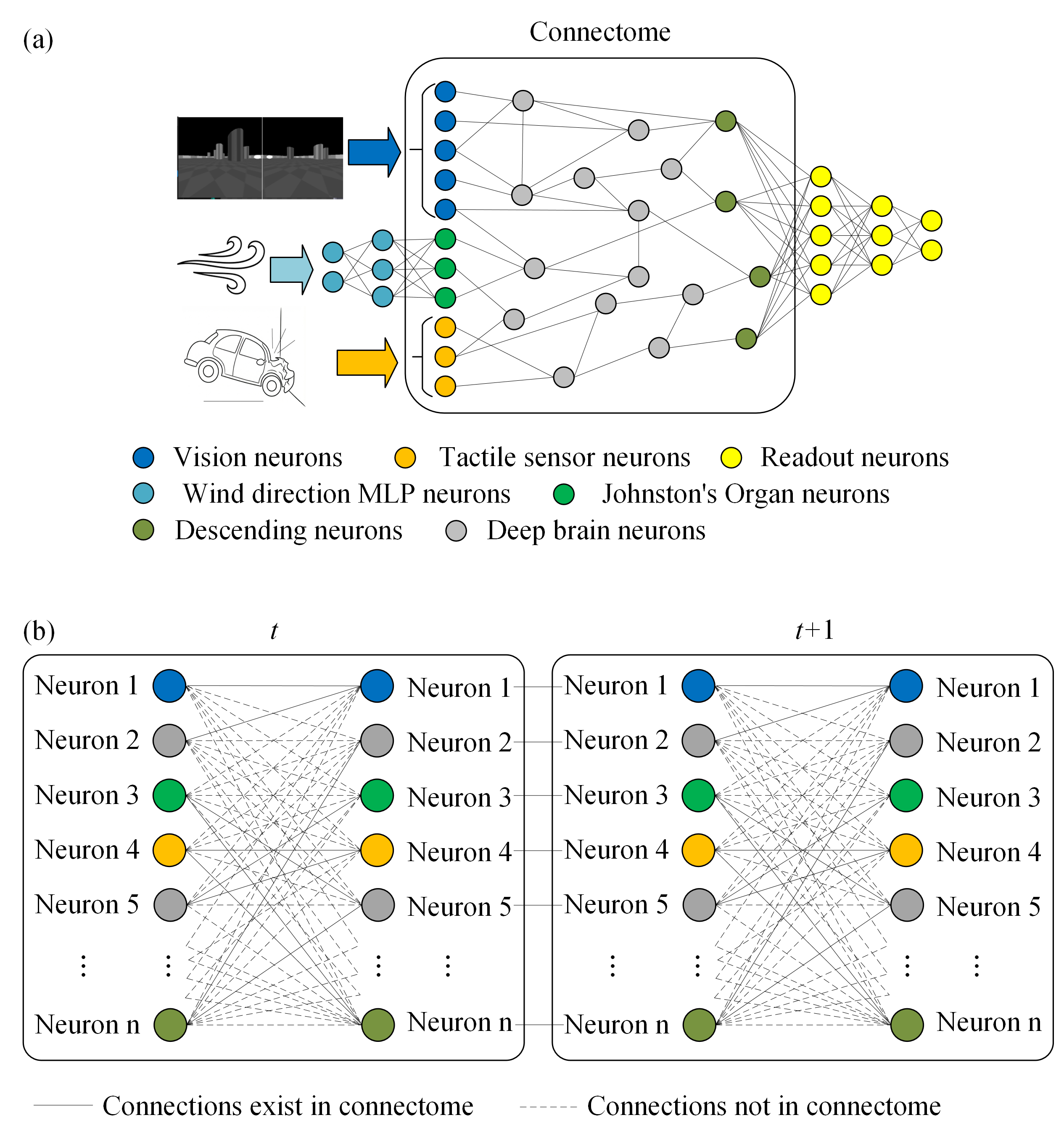}
    \caption{Structure of the FLYNN. (a) Diagram showing the connectome network and its inputs and outputs. (b) Diagram showing the RNN built from the connectome. Sensor input, bias, and activation function were not included for simplicity.}
    \label{fig:network}
\end{figure}

At the core of FLYNN is an RNN constructed from all connectome connections of a \textit{Drosophila} brain. Sensory inputs were passed into corresponding sensory neurons, and motion commands were extracted from descending neurons via a multilayer perceptron (MLP). The sensory and descending neurons used in this work are illustrated in Fig. \ref{fig:neurons}, and the overall structure of FLYNN is shown in Fig. \ref{fig:network}.

\subsection{From Connectome to RNN}

The \textit{Drosophila} connectome dataset was released in 2024 \cite{fly_connectome_1,fly_connectome_2} and is publicly accessible via the FlyWire.ai platform. This is a seminal result in biology in which the neuronal wiring diagram of the complete brain of an advanced organism was precisely mapped by acquiring and analyzing brain images obtained via electron microscopy. Our work utilizes the ``FAFB v783" version of the connectome, derived from a female adult \textit{Drosophila}, and contained all the connections with no less than five synapses. The entire connectome comprised 139,255 neurons and 5,342,445 connections, and was structured as an edge list, where each entry specifies the pre- and post- synaptic neuron IDs, and the corresponding synapse count.

We employed a simple leaky integrator neuron model to construct the network. Specifically, the hidden state $h$ of the neurons at time $t+1$ is defined by the following recurrence relation:

\begin{equation}
  h^{t+1} = \alpha \odot h^t + (1 - \alpha)\odot \tanh(W h^t +  x^t + b) 
\end{equation}


\noindent Here $\odot$ denotes element-wise multiplication, $W$ represents the $n\times n$ synaptic weight matrix where $n$ is the number of neurons, and $h^t$ denotes the states of all neurons at time $t$. The vector $x^t$ represents sensory inputs. Because only a small subset of neurons receives environmental stimuli, most elements of $x^t$ are zero. Finally, $b$ and $\alpha$ represent the biases and leak rates of the neurons, respectively. Here $c(i)$ denotes the class of neuron $i$, so that $\alpha_i = \alpha_{c(i)}$ enforces the constraint that all neurons of the same class share a single leak rate. The class of each neuron is available in the connectome dataset.

At its core, the FLYNN is a biologically-constrained recurrent architecture representing the full FAFB v783 \textit{Drosophila} connectome. Unlike conventional multilayered networks, this network utilizes a single hidden state vector $h^t$ that evolves according to a synaptic connectivity matrix $W$ directly derived from the connectome. Since most possible neural connections in $W$ do not exist in the connectome, the matrix $W$ is characterized by extreme topological sparsity (99.9725\%), where nonzero weights are assigned only to verified biological connections, as illustrated in Fig.~\ref{fig:network}(b). In this configuration, network depth is an emergent property of temporal propagation and integration: as discrete time-steps progress, signals propagate from sensory neurons through intermediate deep brain circuits to descending motor neurons. By representing the entire connectome as a sparse matrix, we preserved all complex, nonhierarchical connectome structures, such as the ring attractor \cite{ring_neuron_2020,ring_neuron_2022}, that are essential for stable heading and navigation. 

\subsection{Input and Output}

To enable effective navigation, we integrated three essential sensory modalities—vision, wind direction, and tactile information—into the network, allowing the robot to see obstacles, orient toward the target, and detect collisions. 

Visual input was provided via images captured from two simulated onboard cameras of the robot. While visual data could theoretically be mapped to the \textit{Drosophila} photoreceptors, incomplete connectivity data for approximately half of the photoreceptors in the left eye rendered this approach infeasible. Instead, we bypassed the photoreceptor layer and mapped visual inputs directly to photoreceptors' primary downstream targets: the L1, L2, and L3 neurons \cite{drosophila_visual_system}. Camera images were first spatially sampled according to the anatomical positions of these L neurons, and the resulting intensity values were then transmitted to respective L neurons. The anatomical positions of L neurons are available in the connectome dataset. Because L1 and L2 neurons primarily detect motion edges, their inputs underwent temporal filtering to enhance edge contrast before being integrated into their hidden states. Conversely, L3 neurons, which process slow fluctuations in environmental luminance, received inputs that were averaged over time prior to integration.

\textit{Drosophila} utilizes wind direction as a critical cue to locate food. They move against the wind upon detecting food odors, a behavior known as  anemotaxis \cite{anemotaxis}. Thus, we used wind direction to represent the relative angle from the robot to the target. Wind direction is primarily sensed through the Johnston’s Organ (JO) located in the antennae \cite{JO_drosophila}. While target angle information should ideally be routed directly to JO neurons, the architectural complexity of JO neurons necessitated an intermediate step. Target angle was first processed by a small MLP, the output of which was then mapped to the JO neurons within the connectome.

In addition to wind sensing, \textit{Drosophila} uses bristle neurons to collect tactile information for collision detection and recovery. Collision data were mapped to specific bristle neurons in the connectome. While bristles are distributed across the entire body, we simplified the input by targeting only those located on the anterior portion of the head. Specifically, a collision on the left side of the head activated all corresponding left-side head bristle neurons. The same logic applied to collisions occurring on the right side of the head. 

The descending neurons of \textit{Drosophila} convey motor commands. A small MLP was connected to all descending neurons in the connectome to convert the encoded hidden states into final wheel commands for the robot.

\subsection{MuJoCo Environment}

\begin{figure}
    \centering
    \includegraphics[width=0.48\textwidth]{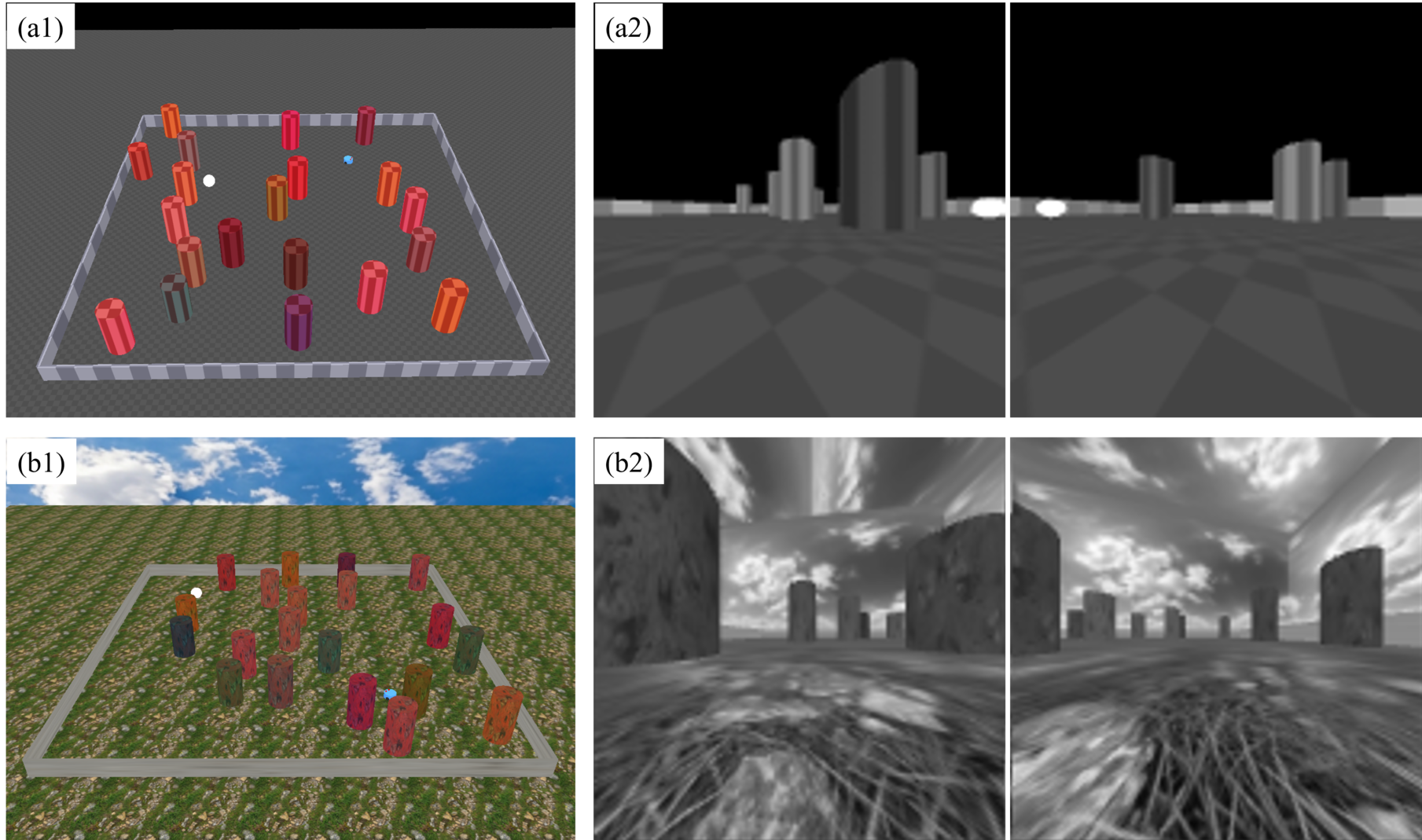}
    \caption{The simulated MuJoCo environments. The cylinders are obstacles, the white ball is the target, and the short blue cylinder is the robot. (a1) 3D rendering of the environment with checkerboard textures in which the networks were trained and evaluated for vision loss robustness.  (a2) The checkerboard textured environment as seen by the left and right cameras on the robot. Camera resolution was set to 128 x 128 pixels. (b1) 3D rendering of the environment in which the networks were evaluated for out-of-distribution generalization. Realistic textures were applied to obstacles, walls, floor, and sky. Note that no model was trained in this environment. (b2) The realistic textured environment as seen by the left and right cameras on the robot.}
    \label{fig:environment}
\end{figure}

We used FLYNN to control a simulated differential drive robot within the environment simulated by the MuJoCo physics engine \cite{2012MuJoCo}. The simulation environment used for training the networks and evaluating their robustness to vision loss is depicted in Fig.~\ref{fig:environment}(a1). The walls enclosed a 14 × 14 m arena populated by 20 randomly distributed cylindrical obstacles. These obstacles had a uniform radius of 0.4 m and were assigned randomized colors. The obstacles, walls, and the floor were textured with checkerboard patterns. The robot was equipped with two grayscale cameras (i.e., eyes), each featuring a 120° field of view (FOV) with a 10° overlap between the two camera FOVs. Representative images as captured by these cameras are illustrated in Fig.~\ref{fig:environment}(a2). Images from the cameras were processed and mapped to the network's left- and right-side L neurons. The agent’s objective was to locate the "food" target, represented by a white sphere randomly placed within the arena. The target's relative angle to the robot was supplied to FLYNN as wind direction, while tactile feedback was provided whenever a collision occurred.

We also modified the environment to evaluate the OOD generalization of different models by replacing the checkerboard textures with photo-realistic textures. 3D rendering and camera images of the modified environment are shown in Figs. \ref{fig:environment}(b1) and (b2). The floor was textured with an image of grassland, the sky was textured with an image of a clear sky, and the obstacles and walls were textured with an image of a concrete surface. This modified environment was used only for OOD generalization evaluation and no model ever saw this environment during training.

\subsection{Model Training}

In this study, we used Dataset Aggregation (DAgger) imitation learning \cite{dagger} to train the FLYNN model to solve navigation tasks in our simulated environment. Imitation learning mirrors how organisms learn from prior successful experience. All model parameters were trainable, including the weights, biases, and per-neuron class leak rates of FLYNN. The expert teacher had access to privileged environmental information, including obstacle coordinates, the target location, and the robot's current pose, while the student only had access to sensory inputs comprising vision, wind direction, and collision signals. The teacher used privileged information to generate optimal paths. These paths were planned using the Vector Field Histogram Star (VFH*) algorithm \cite{vfh_star}, a locally deterministic method that yields smooth, obstacle-avoiding trajectories resembling \textit{Drosophila} navigation and provides stable, consistent training data. A Proportional-Integral-Derivative (PID) controller then translated these paths into target linear and turning velocities, which were converted into differential wheel speeds.

Training data generated by the teacher were segmented into five behavioral categories: start, pre-collision, collision, straight, and turn. This categorization allowed for controlled sampling, ensuring balanced training datasets. The model underwent four training iterations. At the beginning of each training iteration, 500 new episodes were collected, processed, and added to the cumulative training pool. To ensure generalization, obstacle positions and colors, target location, and initial robot pose were randomized for every episode. In the initial training iteration, the robot was controlled exclusively by the expert teacher, corresponding to a teacher-student mixing ratio of 1.0. To enhance data diversity, random noise with the average amplitude of 0.5$\times$ max wheel speed was injected into the teacher's control commands for 50 consecutive steps every 60 steps. Starting with the second iteration, the teacher-student mixing ratio was decreased by 0.5 per iteration, and the injected noise level was reduced by 0.2 per iteration. Each iteration consisted of 300 training steps, with each step comprising 128 batches of data. We employed the Adam optimizer to minimize the Mean Squared Error (MSE) between the teacher's expert commands and the student's outputs.

\subsection{Reference Networks}
\label{sec:reference_networks}
We compared the performance of FLYNN against reference networks described below. To ensure a fair comparison, all reference networks received exactly the same set of sensory inputs, and were trained mostly in the same way with differences described below. 
\subsubsection{EfficientNet and MobileNet}

We used  EfficientNet-B0~\cite{efficientnet} and MobileNet-v3-large \cite{mobilenet} as baseline models for performance comparison. For brevity, these networks are hereafter referred to as EfficientNet and MobileNet, respectively. These networks were chosen because they have roughly the same parameter counts as FLYNN ($\sim 5$ million), and are known to have a strong balance between speed, latency, and accuracy at low parameter counts.

Both networks were trained on the same navigation task as the FLYNN using the same DAgger framework. Camera images were down-sampled to $30\times30$ pixels, matching the approximate resolution of the \textit{Drosophila} eye \cite{drosophila_visual_system}. Images from left and right cameras were then concatenated horizontally before being passed to the models. The original classifiers in these networks were replaced with Gated Recurrent Units (GRUs) followed by an MLP to generate robot control commands. Same as in the FLYNN, relative target angle and collision data were provided to the networks as inputs to the GRU.

Both EfficientNet and MobileNet were trained with camera dropout since CNNs are known to be sensitive to degraded or lost image input without special treatment. During every training session, in each batch of training data, 20\% of chunks had the left side camera turned off, 20\% of chunks had the right side camera turned off, and another 20\% of chunks had both cameras turned off. This forced the agents to learn to navigate under partial or total vision loss conditions.  

\subsubsection{Small-World Model}

Animal brains show small-world network topology \cite{small_world_brain_bassett2017small}, and it has been shown that random small-world networks could perform complex tasks effectively \cite{random_network_xie2019exploring}. Thus, we included a random small-world network as a baseline, referred to as SmallWorldNet, that was designed to closely match important properties of the \textit{Drosophila} connectome.

We utilized the Watts-Strogatz (WS) model \cite{small_world} to generate a graph with 139,255 nodes and an average degree of 78, matching the total synapse count and sparsity of the connectome dataset. Also, to ensure information propagation geometry was preserved, sensory populations were sampled from communities whose average path length to other sensory communities and descending neurons matched that of the connectome. Importantly, SmallWorldNet shares with FLYNN the recurrent leaky-integrator substrate, the per-modality routing of sensory inputs into dedicated neuron populations, the small-world property, the node degree, and the sensory-to-descending path-length statistics; the two networks differ only in the specific wiring, which is taken from the \textit{Drosophila} connectome for FLYNN and generated synthetically for SmallWorldNet. 

The FLYNN--SmallWorldNet comparison serves as an architecture-matched control that isolates the contribution of connectome-specific topology, whereas the comparison against EfficientNet and MobileNet establishes performance parity against standard hand-crafted networks at an equal parameter budget. More details on each network could be found in the source code repository.

\section{Results}

\begin{figure}
    \centering
    \includegraphics[width=0.48\textwidth]{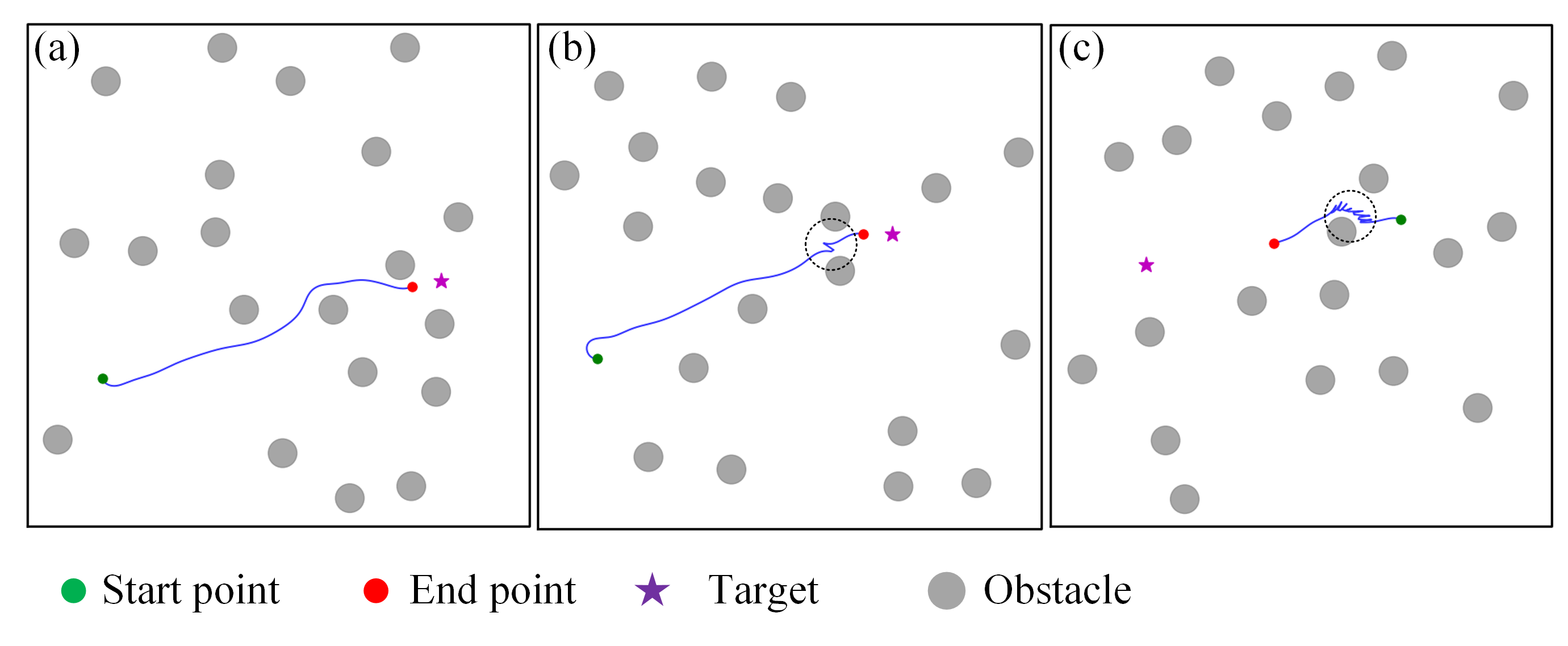}
    \caption{Representative trajectories of the robot controlled by FLYNN with full vision in the checkerboard textured environment. (a) A successful trajectory with no collisions. (b) A successful trajectory with one collision (dashed circle), and successful recovery. (c) A failed trajectory with many collisions (dashed circle). The robot eventually recovered from collisions but didn't have enough time to reach the target.}
    \label{fig:example_trajecoties}
\end{figure}

The trained FLYNN exhibited \textit{Drosophila}-like navigation behaviors, including target pursuit, obstacle avoidance, and post-collision recovery (backing up, turning, and moving forward). Representative trajectories of the agent navigating the checkerboard-textured environment toward the target are shown in Fig.~\ref{fig:example_trajecoties}. Panel (a) shows a representative episode in which the robot successfully navigated around the obstacles and reached the target. In Fig.~\ref{fig:example_trajecoties}(b) the robot had one collision but was able to recover by backing up, turning, and driving forward, illustrated by the Z-shaped trajectory in the dashed circle. Fig.~\ref{fig:example_trajecoties}(c) shows a situation in which the robot had repeated collisions, spent too much time recovering from those collisions, and failed to reach the goal within the 600-step limit.

To quantify the performance of the models, each model was used to control the robot for the same set of 300 randomized episodes, in which obstacles were randomly positioned and colored. The robot's initial position and orientation, as well as the target location, were also randomized. The following parameters were measured for quantifying model performance. 
\begin{enumerate}
    \item Success rate: percentage of successful episodes out of the 300 total episodes. An episode was defined as successful if the robot reached the target within 600 steps, regardless of the number of collisions that occurred during the process. 
    \item Success weighted by Path Length (SPL): SPL of episode $i$ is defined as $SPL_i = S_i\frac{L^*_i}{\max(L^*_i, L_i)}$, where $L^*_i$ is the optimal path length calculated by the A* algorithm, and $L_i$ is the actual path length. SPL standard deviation is omitted because the SPLs are bimodal, making it uninformative. 
    \item SPL of successful episodes only.
    \item Episode duration. One episode would end once the robot reached the target, i.e., the center-to-center distance between the robot and the target was less than 0.8 m. Otherwise the robot would run the entire 600 steps allowed, which corresponds to 12 seconds.
    \item Collision count per episode.
    \item Collision count per episode of successful episodes only.
\end{enumerate}

\subsection{OOD Generalization}

\begin{figure}
    \centering
    \includegraphics[width=0.52\linewidth]{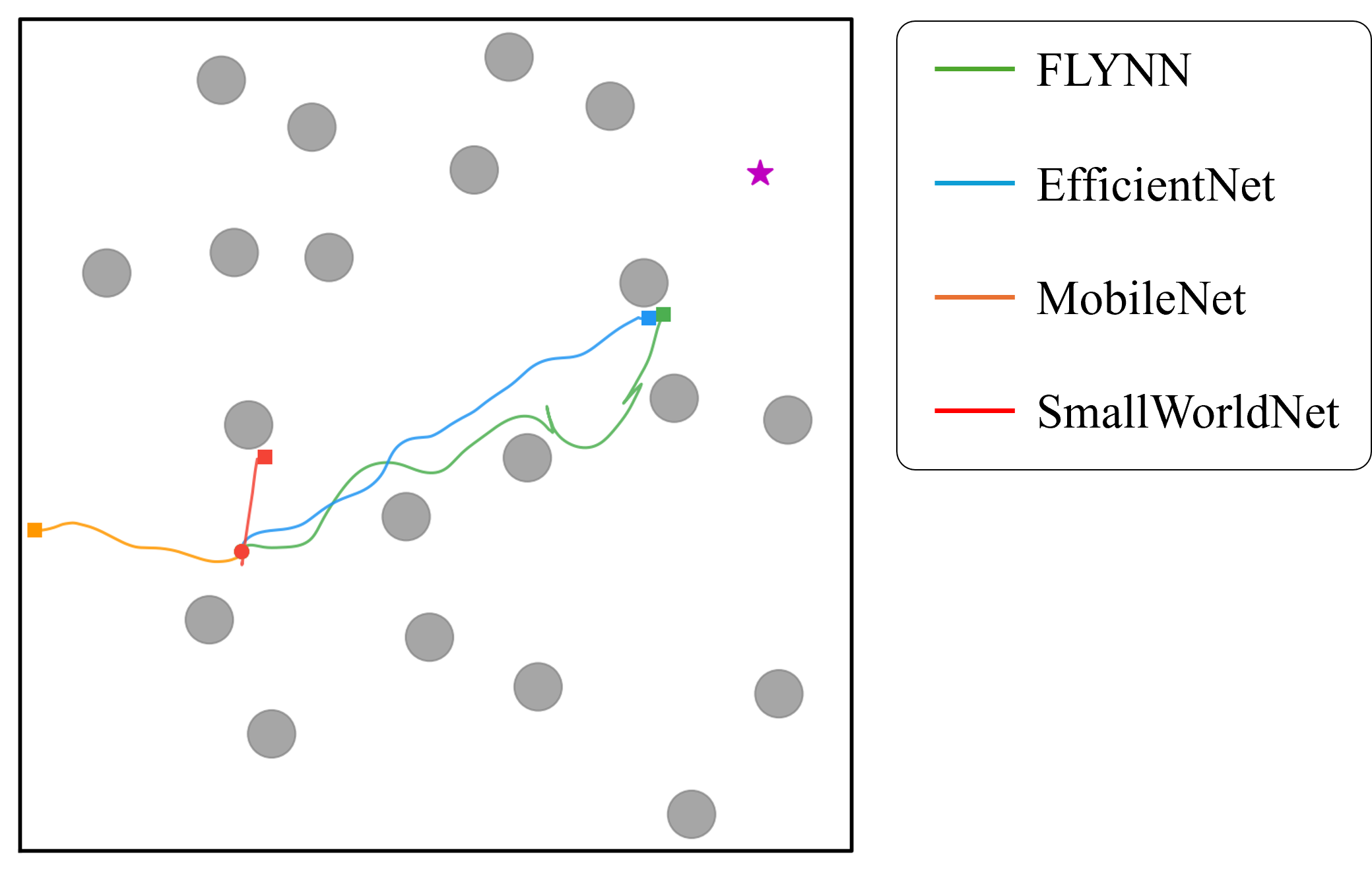}
    \caption{Example trajectories of models driving the robot in the modified environment with realistic textures.}
    \label{fig:trojectory_ood}
\end{figure}

We tested the OOD generalization capability of all the models by having the models control the robot in the modified environment with realistic textures, which none of the models had ever seen during training, while keeping all sensory input intact. Performance parameters are reported in the full-vision rows of Table \ref{tab:performance}. While all models performed comparably in the checkerboard environment, FLYNN showed the best OOD generalization in the modified environment, with the best success rate, average SPL, and average episode duration. It had the highest average collision count both overall and within successful episodes. This is consistent with its maintained collision-recovery behavior, which sometimes required multiple attempts, resulting in a higher collision count. The collision count for other models was low because they either partially or entirely lost the collision recovery behavior and would collide with an obstacle and stall. Example trajectories of all models in the modified environment are shown in Fig. \ref{fig:trojectory_ood}. FLYNN showed all expected navigation behaviors. EfficientNet did not back up from the collision and stalled. MobileNet and SmallWorldNet both failed entirely, driving in the wrong direction and eventually stalling.

\begin{table*}[t]
\centering
\caption{Model performance comparisons across environment textures and vision conditions (number of episodes $n=300$; best values in bold). The full-vision rows use the checkerboard (training) and realistic-texture (OOD) environments; the partial- and total-vision-loss conditions use the checkerboard environment with the indicated camera(s) disabled.}
\label{tab:performance}
\small 
\begin{tabular}{ll c c c c c c}
\toprule
\textbf{Condition} & \textbf{Model} & \textbf{Success \%} & \textbf{SPL} & \textbf{SPL$\vert$Succ.} & \textbf{Episode Dur. (s)}& \textbf{Avg. Coll.} & \textbf{Avg. Coll. $\vert$ Succ.} \\ 
\midrule

\multirow{4}{*}{\shortstack[l]{Checkerboard,\\ full vision}}     & FLYNN          & \textbf{92.9} & \textbf{0.83} & 0.89$\pm$0.10 & \textbf{7.2$\pm$2.3}& \textbf{0.89$\pm$1.64} & \textbf{0.54$\pm$0.97} \\
                                 & EfficientNet  & 90.3 & 0.80 & 0.88$\pm$0.11 & 7.6$\pm$2.4& 1.08$\pm$1.57 & 0.71$\pm$1.05 \\
                                 & MobileNet     & 91.6 & 0.81 & 0.89$\pm$0.11 & 7.4$\pm$2.4& 1.00$\pm$1.46 & 0.67$\pm$0.95 \\
                                 & SmallWorldNet & 89.0 & 0.80 & \textbf{0.90$\pm$0.10} & 7.8$\pm$2.4& 1.12$\pm$2.08 & 0.57$\pm$1.11 \\ 
\midrule

\multirow{4}{*}{\shortstack[l]{Realistic image\\ (OOD), full vision}}& FLYNN          & \textbf{42.1}& \textbf{0.32}& 0.75$\pm$0.09& \textbf{10.6$\pm$1.9}& 3.36$\pm$1.96& 1.53$\pm$0.99\\
                                 & EfficientNet  & 17.2& 0.15&\textbf{ 0.90$\pm$0.09}& 11.3$\pm$1.8& 1.32$\pm$1.32& 0.38$\pm$0.74\\
                                 & MobileNet     & 0.0& 0.0& N.A.& 12.0$\pm$0.0& 0.65$\pm$0.67$^{\dagger}$& N.A.\\
                                 & SmallWorldNet & 3.2& 0.02& 0.67$\pm$0.04& 12.0$\pm$0.2& 1.79$\pm$1.48& 0.2$\pm$0.63$^{\dagger}$\\ 
\midrule

\multirow{4}{*}{Left eye only}   & FLYNN          & 77.4 & 0.67 & 0.87$\pm$0.11 & 8.5$\pm$2.7& 2.27$\pm$2.51 & 1.21$\pm$1.65 \\
                                 & EfficientNet  & \textbf{78.6} & \textbf{0.70} & \textbf{0.89$\pm$0.10} & \textbf{8.2$\pm$2.7}& 1.11$\pm$1.54 & 0.64$\pm$0.91 \\
                                 & MobileNet     & 73.5 & 0.65 & 0.88$\pm$0.10 & 8.3$\pm$2.8& 1.11$\pm$1.46 & \textbf{0.51$\pm$0.76} \\
                                 & SmallWorldNet & 46.3 & 0.40 & 0.86$\pm$0.14 & 10.2$\pm$2.4& \textbf{0.85$\pm$1.17} & 0.57$\pm$0.80 \\ 
\midrule

\multirow{4}{*}{Right eye only}  & FLYNN          & \textbf{69.6} & \textbf{0.60} & 0.86$\pm$0.12 & \textbf{8.7$\pm$2.8}& 2.46$\pm$2.75 & 0.92$\pm$1.42 \\
                                 & EfficientNet  & 46.6 & 0.42 & 0.90$\pm$0.10 & 9.7$\pm$2.8& 1.77$\pm$1.81 & 0.49$\pm$0.84 \\
                                 & MobileNet     & 58.3 & 0.53 & \textbf{0.91$\pm$0.10} & 9.0$\pm$2.9& \textbf{1.22$\pm$1.85} & \textbf{0.31$\pm$0.73} \\
                                 & SmallWorldNet & 68.0 & 0.56 & 0.82$\pm$0.12 & 9.3$\pm$2.4& 1.38$\pm$1.43 & 0.66$\pm$0.80 \\ 
\midrule

\multirow{4}{*}{Total blindness} & FLYNN          & \textbf{44.3} & \textbf{0.37} & 0.83$\pm$0.13 & \textbf{10.3$\pm$2.5}& 4.54$\pm$2.88 & 1.95$\pm$1.97 \\
                                 & EfficientNet  & 3.9  & 0.03 & 0.90$\pm$0.10 & 11.8$\pm$1.0& 1.00$\pm$0.27 & 0.00$\pm$0.00$^{\dagger}$ \\
                                 & MobileNet     & 16.5 & 0.16 & \textbf{0.96$\pm$0.05} & 11.3$\pm$1.8& 1.18$\pm$0.79 & 0.88$\pm$0.93 \\
                                 & SmallWorldNet & 2.6  & 0.02 & 0.75$\pm$0.08 & 11.9$\pm$0.6& 0.72$\pm$1.01$^{\dagger}$ & 1.63$\pm$0.74 \\ 
\bottomrule
\end{tabular}
\\[2pt]
{\footnotesize $^{\dagger}$Low collision counts for models with near-zero success rates reflect stalling rather than effective navigation, and are therefore not highlighted in bold.}
\end{table*}

\subsection{Vision loss tolerance}

\begin{figure}
    \centering
    \includegraphics[width=0.48\textwidth]{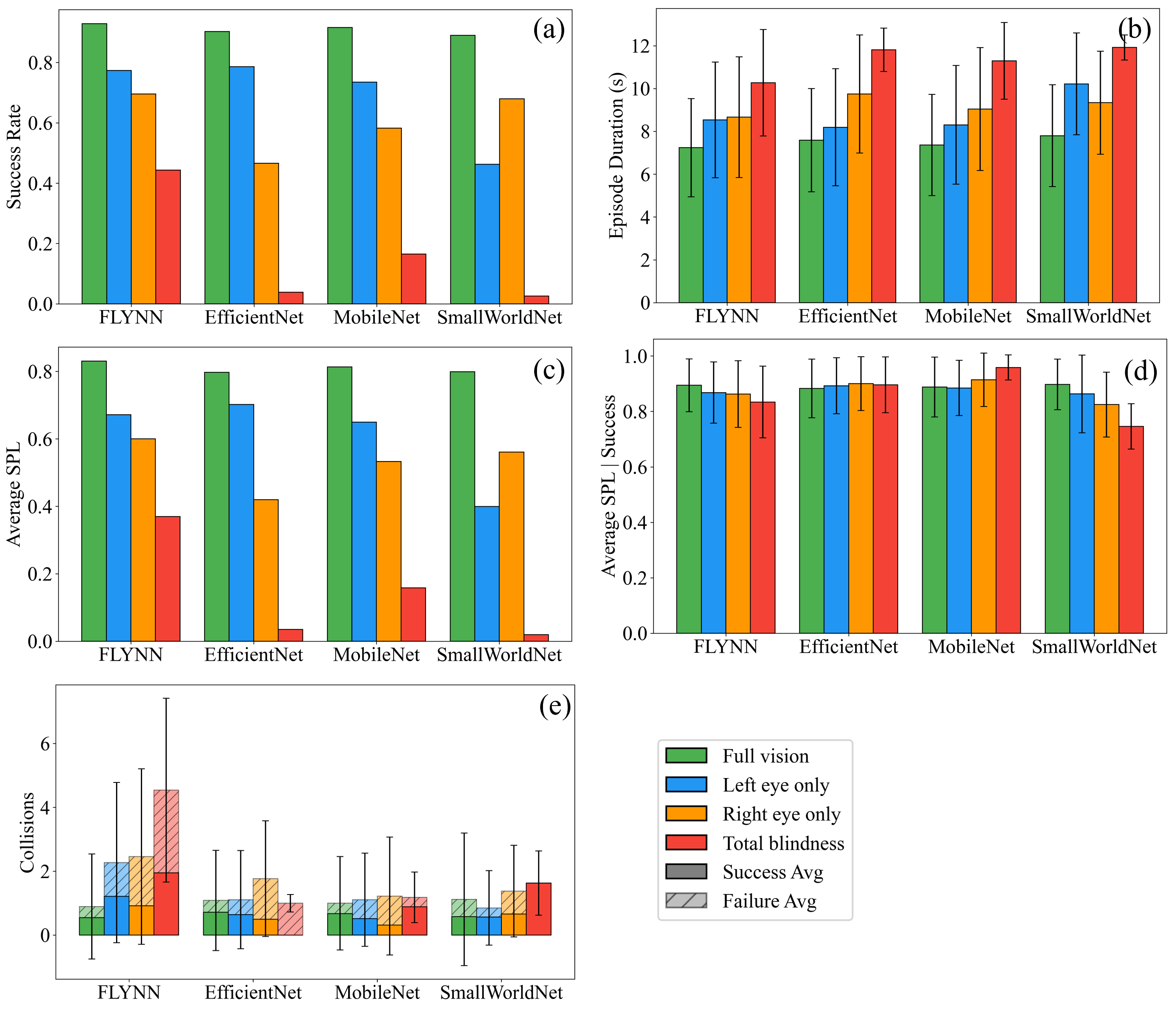}
    \caption{Performance comparisons of different models under different vision conditions. Number of episodes was 300. (a) Success rate. (b) Average episode duration. (c) Average SPL. (d) Average SPL of successful episodes. (e) Number of collisions per episode. Shadow areas show collision count from failed episodes.}
    \label{fig:performance}
\end{figure}

\begin{figure}
    \centering
    \includegraphics[width=0.32\textwidth]{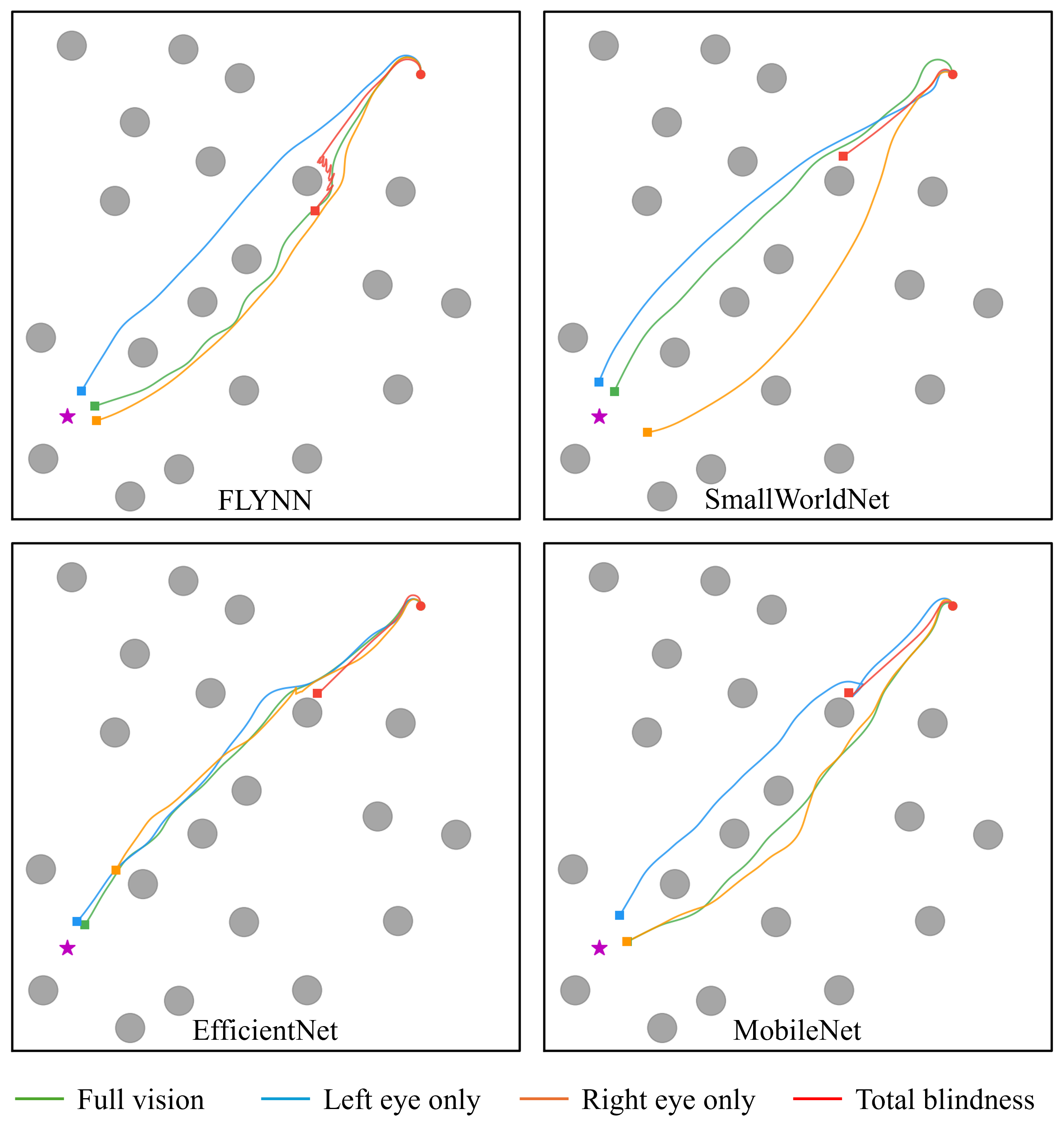}
    \caption{Example trajectories of models under different vision conditions.}
    \label{fig:trajectory_vision_loss}
\end{figure}

We also tested model performance under conditions of partial or total vision loss. This evaluation was conducted in the environment with checkerboard textures, the environment in which the models were trained. Performance parameters of all models are reported in the vision-loss rows of Table \ref{tab:performance}, and also plotted in Fig. \ref{fig:performance} for easy comparison. All models achieved similar performance across all measured metrics with intact vision input, as shown by the green bars in Fig. \ref{fig:performance} and the ``Checkerboard, full vision'' rows of Table \ref{tab:performance}. Visual inspection of navigation episodes confirmed that all models exhibited the expected navigation behaviors (example trajectories are shown in Fig. \ref{fig:trajectory_vision_loss}).

All models showed similar levels of performance degradation when one eye's vision input was removed as shown by blue and orange bars in Fig. \ref{fig:performance}. However, note that  EfficientNet and MobileNet achieved such robustness through the camera dropout training, whereas FLYNN and SmallWorldNet demonstrated this robustness spontaneously without camera dropout training. We observed that EfficientNet and MobileNet would fail entirely without dropout training, i.e., success rates were 0\%.

FLYNN showed the highest robustness when vision input from both eyes was lost. As shown by the red bars in Fig. \ref{fig:performance}(a) and (c), success rate and average SPL collapsed for all networks under total vision loss, except for FLYNN. Since the wind-direction (target-angle) and collision cues were available to all models, the divergence under total blindness reflects a difference in the ability to exploit these residual signals rather than a difference in the signals themselves. The high success rate of FLYNN suggests that FLYNN was able to fall back on the remaining sensory inputs, i.e., wind direction and collision information, to maintain its navigational capability. The low episode duration, as shown in Fig. \ref{fig:performance}(b), suggests that the robot rarely stalled. This agrees with the observed high collision count (Fig. \ref{fig:performance}(e)). The robot maintained collision recovery behavior, but without the help of vision information, it took more attempts to recover (Fig. \ref{fig:trajectory_vision_loss}, red trajectory of FLYNN). In contrast, as shown by the red trajectories in Fig. \ref{fig:trajectory_vision_loss}, all models except FLYNN stalled once a collision happened, i.e., collision recovery behavior was lost under total blindness. This agrees with the relatively low collision count as shown in Fig. \ref{fig:performance}(e).  

The high average SPLs of successful episodes across all models and all conditions, as shown in Fig. \ref{fig:performance}(d), suggests that no model drove the robot along spiral or erratic paths in successful episodes. However, the success rate must be taken into consideration when evaluating SPLs. For example, in the rare situation where there is no obstacle between the robot and the target, the episode could turn into a successfully completed episode with high SPL due to a direct path. The high average success SPL thus merely indicates that the models were driving mostly straight.

\begin{figure}
    \centering
    \includegraphics[width=0.48\textwidth]{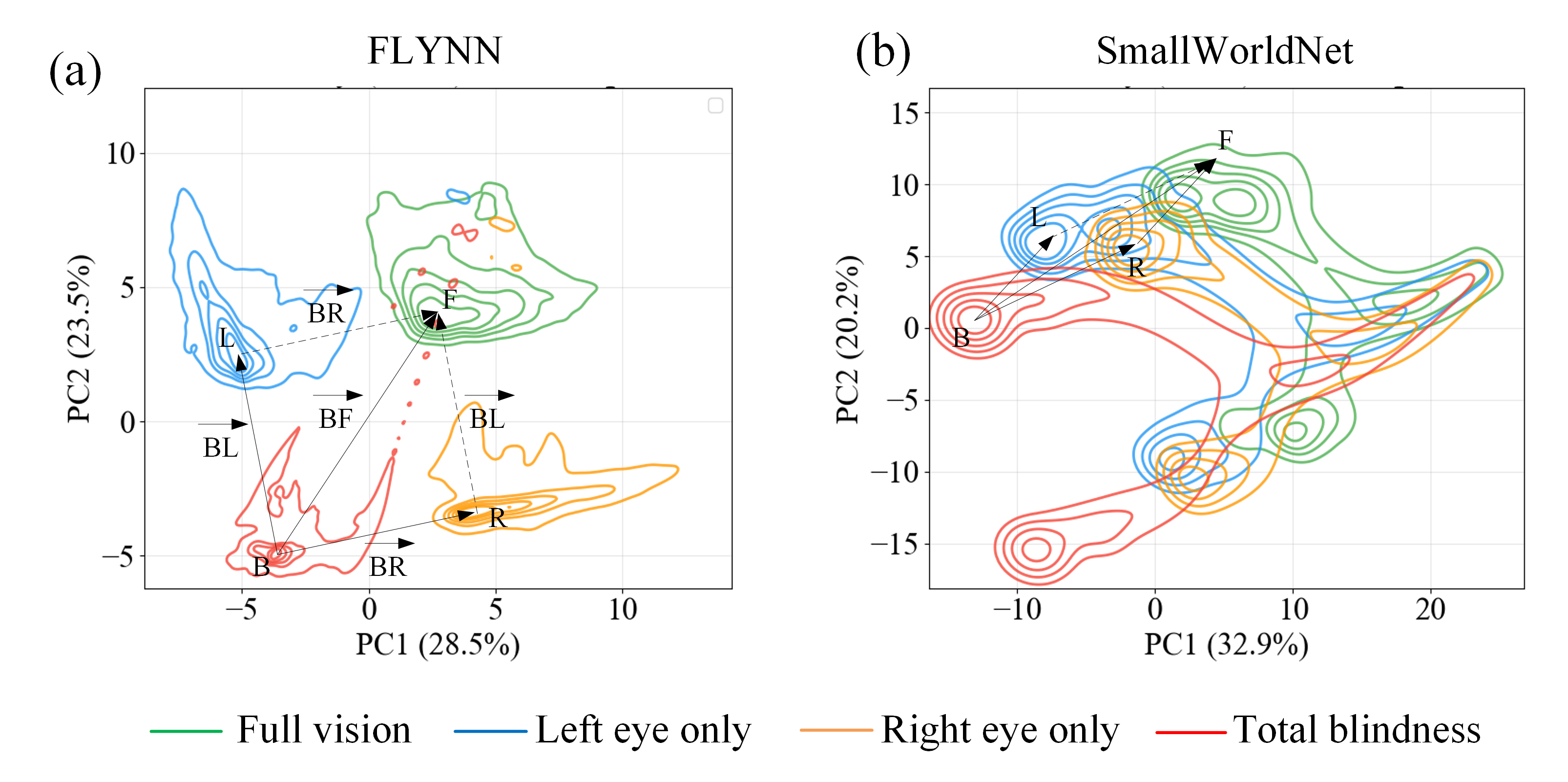}
    \caption{KDE plots of the first two principal components of (a) FLYNN's and (b) SmallWorldNet's internal states under different vision conditions over 300 episodes.}
    \label{fig:pca_analysis}
\end{figure}

Notice that similar to FLYNN, SmallWorldNet also showed some resistance to partial vision loss without dropout training. However, it failed almost entirely with total vision loss, whereas FLYNN remained functional.  Since both networks shared similar properties -- the RNN architecture, small-world property, node degree, distances between sensory neurons and descending neurons -- looking into how their internal states differ could provide insights into the source of FLYNN's robustness. 

We performed Principal Component Analysis (PCA) on the neuron states of both models as they navigated through the same set of 300 test environments. We then performed Kernel Density Estimation (KDE) on the first two principal components of all the PCA trajectories over 300 episodes. The KDE analysis showed where the models' internal states appear most often in the principal component space when navigating through the episodes. The results are plotted in Fig. \ref{fig:pca_analysis}. It was evident that the internal states of FLYNN formed distinct clusters under different vision conditions, indicating a clear and stable internal representation for each vision condition. In contrast, SmallWorldNet produced broad, overlapping distributions with high representational entropy.

We found that in FLYNN the sum of the vectors $\vec{BL}$ and $\vec{BR}$ -- representing the transitions from KDE peak of the totally blind state $B$ to single-eye states $L$ and $R$ -- almost exactly equal to the blind ($B$) to full-vision ($F$) states vector $\vec{BF}$ , i.e., $\vec{BF} \approx \vec{BL} + \vec{BR}$ (arrows in Fig. \ref{fig:pca_analysis}(a)). We further numerically checked the relationship between the vectors using the first 10 principal components, which together explained 89.1\% total variance for FLYNN and 94.6\% for SmallWorldNet. For FLYNN, vector magnitudes were$\vert\vec{BF} \vert=11.16$, $\vert \vec{BL} + \vec{BR} \vert=11.11$, and cosine similarity was 0.9998. For SmallWorldNet, $\vert\vec{BF} \vert=22.88$, $\vert \vec{BL} + \vec{BR} \vert=30.01$, and cosine similarity was 0.9439. The linear sum held almost perfectly for FLYNN and did not hold for SmallWorldNet. 

This linear superposition of FLYNN internal states indicates that the \textit{Drosophila} connectome preserves the independent information streams of each sensor, allowing the agent to maintain a coherent navigation logic even when the vision input is partially or entirely removed. Such geometric regularity was absent from the small-world network baseline. Note that this analysis is only correlational: it characterizes the geometry of the learned internal states rather than establishing a causal mechanism.

The OOD robustness and the vision-loss robustness of FLYNN need not share a single mechanism. The temporal filtering we did for L1 and L2 neuron inputs emphasized motion edges and could partly account for FLYNN's OOD robustness. However note that the temporal filtering function is intrinsic to the L neurons and here the function was implemented manually simply because of the incompleteness of the connectome. The total-vision-loss results, where the visual pathway is removed entirely, shows a more direct contribution of the connectome architecture to FLYNN's vision-loss robustness.

\section{CONCLUSIONS}


In conclusion, this research represents the first effort to build an RNN with a network structure strictly constrained by the \textit{Drosophila} connectome topology. We demonstrated that such a network not only could be trained for complex tasks, such as vision-based multi-sensory navigation, but also achieve performance comparable to modern hand-crafted networks. More importantly, such a network showed remarkable OOD generalization and tolerance to vision loss even though the network never encountered such situations during training. The network was able to maintain most of its navigation capability under unfamiliar conditions, outperforming modern hand-crafted networks in simulated navigation tasks.

The fact that SmallWorldNet showed some degree of robustness to partial vision loss suggests that the small-world property, which is shared between SmallWorldNet and FLYNN, may contribute to the observed robustness. However, the small-world property could not explain all of the robustness that FLYNN exhibited. Our KDE analysis of the PCA-projected internal states revealed distinct differences in the internal state distribution between the two networks. The tight clustering observed in the KDE plots and the approximate linear summation of different states are consistent with the \textit{Drosophila} connectome acting as a topological prior for linear sensory integration. This may allow the agent to gracefully degrade in response to sensor loss without disrupting the global navigation manifold, a property that we did not observe in the random small-world architecture. Such linear sensory fusion may stem from the modular design of the \textit{Drosophila} brain, and could be a potential explanation for the robustness of FLYNN. However, this analysis is geometric and correlational, identifying which specific topological properties drive this behavior remains an important direction for future work.

We acknowledge that the FLYNN built in this work is constrained solely by connectome topology, and we used several simplifications to make constructing a functional network from the complex connectome feasible. Biological properties omitted here include spiking neurons, electrical synapses, gap junctions, axon-length differences, and neurotransmitter types; incorporating any of these could potentially improve performance and bring the model closer to real \textit{Drosophila} behavior. We also note that the OOD evaluation in this work is limited to a visual domain shift: the realistic-texture environment changes low-level image appearance while keeping the arena geometry, obstacle statistics, and task structure unchanged. Robustness to broader distribution shifts, such as changes in dynamics and sensor noise, remains to be evaluated, as does validation on a physical robot.

Overall, we demonstrated both the feasibility of training FLYNN for navigation and its superior robustness to OOD data and input deprivation. We envision this work as a blueprint for designing robust, resilient neural networks and as a platform for studying how biological brains achieve such resilience.



\section*{ACKNOWLEDGMENT}
The work reported herein was supported by the National Science Foundation (NSF) (Award \#IIS-2440583). Any opinions, findings, conclusions or recommendations expressed in this material are those of the authors and do not necessarily reflect the views of the NSF.

The authors acknowledge the use of Google Gemini to improve the language, grammar, and clarity of this paper.

\bibliographystyle{IEEEtran}
\bibliography{citations}

@article{2024Lappalainen,
  title={Connectome-constrained networks predict neural activity across the fly visual system},
  author={Lappalainen, Janne K and Tschopp, Fabian D and Prakhya, Sridhama and McGill, Mason and Nern, Aljoscha and Shinomiya, Kazunori and Takemura, Shin-ya and Gruntman, Eyal and Macke, Jakob H and Turaga, Srinivas C},
  journal={Nature},
  volume={634},
  number={8036},
  pages={1132--1140},
  year={2024},
  publisher={Nature Publishing Group UK London}
}

@article{2024Shiu,
  title={A Drosophila computational brain model reveals sensorimotor processing},
  author={Shiu, Philip K and Sterne, Gabriella R and Spiller, Nico and Franconville, Romain and Sandoval, Andrea and Zhou, Joie and Simha, Neha and Kang, Chan Hyuk and Yu, Seongbong and Kim, Jinseop S and others},
  journal={Nature},
  volume={634},
  number={8032},
  pages={210--219},
  year={2024},
  publisher={Nature Publishing Group UK London},
  doi = {10.1038/s41586-024-07763-9},
  URL = {https://www.nature.com/articles/s41586-024-07763-9}
}

@INPROCEEDINGS{2012MuJoCo,
  author={Todorov, Emanuel and Erez, Tom and Tassa, Yuval},
  booktitle={2012 IEEE/RSJ International Conference on Intelligent Robots and Systems}, 
  title={MuJoCo: A physics engine for model-based control}, 
  year={2012},
  volume={},
  number={},
  pages={5026-5033},
  doi={10.1109/IROS.2012.6386109}}

@article{perceptron1958,
  title={The perceptron: a probabilistic model for information storage and organization in the brain.},
  author={Rosenblatt, Frank},
  journal={Psychological review},
  volume={65},
  number={6},
  pages={386},
  year={1958},
  publisher={American Psychological Association}
}

@inproceedings{CNN_AlexNet,
 author = {Krizhevsky, Alex and Sutskever, Ilya and Hinton, Geoffrey E},
 booktitle = {Advances in Neural Information Processing Systems},
 editor = {F. Pereira and C.J. Burges and L. Bottou and K.Q. Weinberger},
 pages = {},
 publisher = {Curran Associates, Inc.},
 title = {ImageNet Classification with Deep Convolutional Neural Networks},
 volume = {25},
 year = {2012}
}

@InProceedings{input_corruption_miscov,
    author    = {Przewiezlikowski, Marcin and Smieja, Marek and Struski, Lukasz and Tabor, Jacek},
    title     = {MisConv: Convolutional Neural Networks for Missing Data},
    booktitle = {Proceedings of the IEEE/CVF Winter Conference on Applications of Computer Vision (WACV)},
    month     = {January},
    year      = {2022},
    pages     = {2060-2069}
}

@article{input_occlusion,
title = {Are deep learning models robust to partial object occlusion in visual recognition tasks},
journal = {Pattern Recognition},
volume = {171},
pages = {112215},
year = {2026},
issn = {0031-3203},
doi = {https://doi.org/10.1016/j.patcog.2025.112215},
author = {Kaleb Kassaw and Francesco Luzi and Leslie M. Collins and Jordan M. Malof},
}

@article{input_sensor_failure,
title = {A deep learning gated architecture for UGV navigation robust to sensor failures},
journal = {Robotics and Autonomous Systems},
volume = {116},
pages = {80-97},
year = {2019},
issn = {0921-8890},
doi = {https://doi.org/10.1016/j.robot.2019.03.001},
author = {Naman Patel and Anna Choromanska and Prashanth Krishnamurthy and Farshad Khorrami},
}

@InProceedings{efficientnet,
  title = 	 {{E}fficient{N}et: Rethinking Model Scaling for Convolutional Neural Networks},
  author =       {Tan, Mingxing and Le, Quoc},
  booktitle = 	 {Proceedings of the 36th International Conference on Machine Learning},
  pages = 	 {6105--6114},
  year = 	 {2019},
  editor = 	 {Chaudhuri, Kamalika and Salakhutdinov, Ruslan},
  volume = 	 {97},
  series = 	 {Proceedings of Machine Learning Research},
  month = 	 {09--15 Jun},
  publisher =    {PMLR},

}

@InProceedings{mobilenet,
author = {Howard, Andrew and Sandler, Mark and Chu, Grace and Chen, Liang-Chieh and Chen, Bo and Tan, Mingxing and Wang, Weijun and Zhu, Yukun and Pang, Ruoming and Vasudevan, Vijay and Le, Quoc V. and Adam, Hartwig},
title = {Searching for MobileNetV3},
booktitle = {Proceedings of the IEEE/CVF International Conference on Computer Vision (ICCV)},
month = {October},
year = {2019}
}

@article{fly_connectome_1,
  title={Neuronal wiring diagram of an adult brain},
  author={Dorkenwald, Sven and Matsliah, Arie and Sterling, Amy R and Schlegel, Philipp and Yu, Szi-Chieh and McKellar, Claire E and Lin, Albert and Costa, Marta and Eichler, Katharina and Yin, Yijie and others},
  journal={Nature},
  volume={634},
  number={8032},
  pages={124--138},
  year={2024},
  publisher={Nature Publishing Group UK London}
}

@article{fly_connectome_2,
  title={Whole-brain annotation and multi-connectome cell typing of Drosophila},
  author={Schlegel, Philipp and Yin, Yijie and Bates, Alexander S and Dorkenwald, Sven and Eichler, Katharina and Brooks, Paul and Han, Daniel S and Gkantia, Marina and Dos Santos, Marcia and Munnelly, Eva J and others},
  journal={Nature},
  volume={634},
  number={8032},
  pages={139--152},
  year={2024},
  publisher={Nature Publishing Group UK London}
}

@article{drosophila_visual_system,
  title={The Drosophila visual system: From neural circuits to behavior},
  author={Zhu, Yan},
  journal={Cell adhesion \& migration},
  volume={7},
  number={4},
  pages={333--344},
  year={2013},
  publisher={Taylor & Francis}
}

@article{anemotaxis,
  title={Anemotaxis in Drosophila},
  author={Kalmus, H},
  journal={Nature},
  volume={150},
  number={3805},
  pages={405--405},
  year={1942},
  publisher={Nature Publishing Group UK London}
}

@article{JO_drosophila,
  title={The Drosophila auditory system},
  author={Boekhoff-Falk, Grace and Eberl, Daniel F},
  journal={Wiley Interdisciplinary Reviews: Developmental Biology},
  volume={3},
  number={2},
  pages={179--191},
  year={2014},
  publisher={Wiley Online Library}
}

@article{attention,
  title={Attention is all you need},
  author={Vaswani, Ashish and Shazeer, Noam and Parmar, Niki and Uszkoreit, Jakob and Jones, Llion and Gomez, Aidan N and Kaiser, {\L}ukasz and Polosukhin, Illia},
  journal={Advances in neural information processing systems},
  volume={30},
  year={2017}
}

@article{how_flys_see_motion,
  title={How flies see motion},
  author={Borst, Alexander and Groschner, Lukas N},
  journal={Annual review of neuroscience},
  volume={46},
  number={1},
  pages={17--37},
  year={2023},
  publisher={Annual Reviews}
}

@article{ring_neuron_2022,
  title={Flexible navigational computations in the Drosophila central complex},
  author={Fisher, Yvette E},
  journal={Current opinion in neurobiology},
  volume={73},
  pages={102514},
  year={2022},
  publisher={Elsevier}
}

@article{ring_neuron_2020,
  title={The neuroanatomical ultrastructure and function of a biological ring attractor},
  author={Turner-Evans, Daniel B and Jensen, Kristopher T and Ali, Saba and Paterson, Tyler and Sheridan, Arlo and Ray, Robert P and Wolff, Tanya and Lauritzen, J Scott and Rubin, Gerald M and Bock, Davi D and others},
  journal={Neuron},
  volume={108},
  number={1},
  pages={145--163},
  year={2020},
  publisher={Elsevier}
}

@INPROCEEDINGS{vfh_star,
  author={Ulrich, I. and Borenstein, J.},
  booktitle={Proceedings 2000 ICRA. Millennium Conference. IEEE International Conference on Robotics and Automation. Symposia Proceedings (Cat. No.00CH37065)}, 
  title={VFH*: local obstacle avoidance with look-ahead verification}, 
  year={2000},
  volume={3},
  number={},
  pages={2505-2511 vol.3},
  doi={10.1109/ROBOT.2000.846405}}

@inproceedings{dagger,
  title={A reduction of imitation learning and structured prediction to no-regret online learning},
  author={Ross, St{\'e}phane and Gordon, Geoffrey and Bagnell, Drew},
  booktitle={Proceedings of the fourteenth international conference on artificial intelligence and statistics},
  pages={627--635},
  year={2011},
  organization={JMLR Workshop and Conference Proceedings}
}

@article{small_world,
  title={Collective dynamics of ‘small-world’networks},
  author={Watts, Duncan J and Strogatz, Steven H},
  journal={nature},
  volume={393},
  number={6684},
  pages={440--442},
  year={1998},
  publisher={Nature Publishing Group}
}

@article{lstm,
  title={Long short-term memory},
  author={Hochreiter, Sepp and Schmidhuber, J{\"u}rgen},
  journal={Neural computation},
  volume={9},
  number={8},
  pages={1735--1780},
  year={1997},
  publisher={MIT press}
}

@article{bptt,
  title={Backpropagation through time: what it does and how to do it},
  author={Werbos, Paul J},
  journal={Proceedings of the IEEE},
  volume={78},
  number={10},
  pages={1550--1560},
  year={2002},
  publisher={IEEE}
}

@article{ceccarelli2022rgb,
  title={RGB cameras failures and their effects in autonomous driving applications},
  author={Ceccarelli, Andrea and Secci, Francesco},
  journal={IEEE Transactions on Dependable and Secure Computing},
  volume={20},
  number={4},
  pages={2731--2745},
  year={2022},
  publisher={IEEE}
}

@inproceedings{shoeb2025out,
  title={Out-of-distribution segmentation in autonomous driving: Problems and state of the art},
  author={Shoeb, Youssef and Nowzad, Azarm and Gottschalk, Hanno},
  booktitle={Proceedings of the Computer Vision and Pattern Recognition Conference},
  pages={4310--4320},
  year={2025}
}

@article{drosophila_taste_sys_sim,
  title={Connectomic analysis of taste circuits in Drosophila},
  author={Walker, Sydney R and Pe{\~n}a-Garcia, Marco and Devineni, Anita V},
  journal={Scientific Reports},
  volume={15},
  number={1},
  pages={5278},
  year={2025},
  publisher={Nature Publishing Group UK London}
}

@inproceedings{domain_rand_zhang2024,
  title={Robot Vision-Based Autonomous Navigation Method Using Sim2Real Domain Adaptation},
  author={Zhang, Mengjiao and Duan, Shihong and Xu, Cheng},
  booktitle={2024 IEEE International Symposium on Parallel and Distributed Processing with Applications (ISPA)},
  pages={1203--1209},
  year={2024},
  organization={IEEE}
}

@article{liqid_net,
  title={Robust flight navigation out of distribution with liquid neural networks},
  author={Chahine, Makram and Hasani, Ramin and Kao, Patrick and Ray, Aaron and Shubert, Ryan and Lechner, Mathias and Amini, Alexander and Rus, Daniela},
  journal={Science Robotics},
  volume={8},
  number={77},
  pages={eadc8892},
  year={2023},
  publisher={American Association for the Advancement of Science}
}

@inproceedings{liquidnet_suresh2025liquid,
  title={Liquid Neural Networks for Autonomous Driving: A Framework for Intelligent Decision-Making},
  author={Suresh, Yashaswini and Devashya, Vismaya and others},
  booktitle={2025 5th International Conference on Emerging Research in Electronics, Computer Science and Technology (ICERECT)},
  pages={1--6},
  year={2025},
  organization={IEEE}
}

@inproceedings{llm_nav_qiao2025open,
  title={Open-nav: Exploring zero-shot vision-and-language navigation in continuous environment with open-source llms},
  author={Qiao, Yanyuan and Lyu, Wenqi and Wang, Hui and Wang, Zixu and Li, Zerui and Zhang, Yuan and Tan, Mingkui and Wu, Qi},
  booktitle={2025 IEEE International Conference on Robotics and Automation (ICRA)},
  pages={6710--6717},
  year={2025},
  organization={IEEE}
}

@inproceedings{llm_nav_aasi2025generating,
  title={Generating out-of-distribution scenarios using language models},
  author={Aasi, Erfan and Nguyen, Phat and Sreeram, Shiva and Rosman, Guy and Karaman, Sertac and Rus, Daniela},
  booktitle={2025 IEEE International Conference on Robotics and Automation (ICRA)},
  pages={10616--10623},
  year={2025},
  organization={IEEE}
}

@inproceedings{random_network_xie2019exploring,
  title={Exploring randomly wired neural networks for image recognition},
  author={Xie, Saining and Kirillov, Alexander and Girshick, Ross and He, Kaiming},
  booktitle={Proceedings of the IEEE/CVF international conference on computer vision},
  pages={1284--1293},
  year={2019}
}

@article{small_world_brain_bassett2017small,
  title={Small-world brain networks revisited},
  author={Bassett, Danielle S and Bullmore, Edward T},
  journal={The Neuroscientist},
  volume={23},
  number={5},
  pages={499--516},
  year={2017},
  publisher={Sage Publications Sage CA: Los Angeles, CA}
}

\end{document}